\documentclass[a4paper,12pt]{article}
\usepackage[utf8]{inputenc}
\usepackage[english]{babel}
\usepackage{authblk}
\usepackage{graphicx}
\usepackage{caption}
\usepackage{float}      % optional, for H placement

\DeclareGraphicsExtensions{.pdf,.png,.jpg}
\usepackage{mathptmx}
\usepackage[singlespacing]{setspace}
\usepackage[headheight=1in,margin=1in]{geometry}
\usepackage{fancyhdr}
\usepackage[round,authoryear]{natbib}

\bibpunct{(}{)}{;}{a}{,}{,}
\setcitestyle{authoryear,open={(},close={)},aysep={,},yysep={;}}
\usepackage{adjustbox}
\usepackage{booktabs}
\usepackage{array}
\usepackage{amsmath}
\usepackage{hyperref}

\makeatletter
\def\@maketitle{%
  \newpage

  \begin{center}%
  \let \footnote \thanks
    {\LARGE \@title \par}%
  \end{center}%
  \par
  \vskip 0.1em}
\makeatother

\renewcommand{\sectionmark}[1]{}
\renewcommand{\subsectionmark}[1]{}

\graphicspath{{images/}}

\title{Are generative AI text annotations systematically biased?}

\begin{document}

\begin{titlepage}
  \centering
  {\LARGE\bfseries Are generative AI text annotations systematically biased?\par}
  \vspace{1.25cm}

  {\large
    Sjoerd B. Stolwijk,\textsuperscript{1}\quad
    Mark Boukes,\textsuperscript{2}\quad
    Damian Trilling\textsuperscript{3}\par}
  \vspace{0.8cm}

  \textsuperscript{1}Utrecht University, Utrecht School of Governance (USBO)\\
  \textsuperscript{2}University of Amsterdam\\
  \textsuperscript{3}Vrije Universiteit Amsterdam
  \vfill

  \begin{minipage}{0.85\textwidth}
    \centering
    Corresponding author:\\[0.2cm]
    Sjoerd B. Stolwijk\\
    Utrecht School of Governance (USBO), Utrecht University\\
    Email: \texttt{s.b.stolwijk@uu.nl}
  \end{minipage}

  \vspace{1cm}

  \today

  \vfill
  \thispagestyle{fancy}
\end{titlepage}

\maketitle
\thispagestyle{fancy}

\begin{center}
%Computational Methods, artificial intelligence, discourse analysis, large language models, simulation, text analysis, text mining
\newline
\end{center}

\section*{Extended Abstract}

Generative AI models (GLLM) like openAI's GPT4 are revolutionizing the field of automatic content analysis through impressive performance \citep{gilardi_chatgpt_2023, heseltine_large_2024, tornberg_large_2024}. However, there are also concerns about their potential biases \citep[e.g.][]{ferrara_fairness_2024, motoki_more_2024, fulgu_surprising_2024}. So far, these critiques mainly focus on the answers GLLMs generate in conversations or surveys; yet the same concerns could likely apply to text annotations. If this is the case, the impressive performance of GLLMs reported using traditional performance metrics like $F_1$ scores might give a deceptive impression of the quality of the annotations. This paper will investigate the existence and random versus systematic nature of the GLLM annotation bias and the ability of $F_1$ scores to detect these biases.

Potential GLLM annotation biases are consequential: On the one hand, if each researcher used the same GLLM or different GLLMs are biased in the same direction, their substantive results could be biased in the same direction, making it more difficult for cumulative research to weed out biases in individual papers. Alternatively, if different researchers use different GLLMs and each GLLM yields different -- undetected -- biases, this could lead to contrasting and confusing research results, hampering the progress of the field. On top of this, the effect of prompts used to query the GLLM can be strong and unpredictable \citep{kaddour_challenges_2023, webson_prompt-based_2022}. Recent work by \citet{baumann_large_2025} even suggests that modifying prompts could lead to opposite downstream results.

\subsection*{Design}
This paper conceptually replicates the analysis in \citet{boukes_deliberation_2024}, which uses a manual content analysis to find out whether YouTube replies to satire versus non-satire newsvideo's differ in terms of deliberative quality on a number of indicators (political content, interactivity, rationality, incivility, and ideology). We examine the effect of using various GLLMs (Llama3.1:8b, Llama3.3:70b, GPT4o, Qwen2.5:72b) in combination with five different prompts compared to the manual annotations used in that paper. We selected our prompts by translating the original codebook of \citet{boukes_deliberation_2024} into a prompt (``Boukes'') and asking GPT4o to reformulate it, only changing punctuation (``Simpa1'') or using different words (``Para1'', ``Para2''). We added one more prompt (``Jaidka'') based on the crowd-coding instructions of \citet{jaidka_brevity_2019}\footnote{available for four of our five concepts} to evaluate the effect of different operationalizations used in the literature on the results.

We evaluated our GLLM annotations of the original manually coded sample from \citet{boukes_deliberation_2024} in five ways. \textbf{First}, we computed standard evaluation metrics (accuracy, macro average $F_1$). \textbf{Second}, we considered whether GLLMs might differ from manual annotations in terms of prevalence: the number of YT-replies labeled as positive for the concept. \textbf{Third}, we computed a simplified version of the analysis in that paper: the raw correlation between genre (satire vs. non-satire) and the prevalence of each concept according to the GLLM annotations and compared this to the same correlation based on the manual annotations. \textbf{Fourth}, to investigate whether any bias is random or systematic, we calculated the commonality between the different GLLM annotations and their overlap with the manual annotations, by comparing them to four sets of simulated annotations. \textbf{Fifth}, we analyzed the relation between GLLM bias and $F_1$ score. Due to space constraints, we only show results here for the concept of rationality, which most clearly illustrates our findings.

\subsection*{Results}

\begin{table}[H]
    \centering
    \caption{Performance in terms of macro average $F_1$ and accuracy of each prompt-model combination in classifying rationality $N$ = 2459.}
    \label{tab:perf_rationality}
    \begin{tabular}{|l|l|l|l|}
        \hline
        \textbf{GLLM} & \textbf{Prompt} & \textbf{Macro $F_1$} & \textbf{Accuracy} \\ \hline
        gpt4o & Boukes & 0.63 & 0.85 \\ \hline
        gpt4o & Jaidka & 0.69 & 0.85 \\ \hline
        gpt4o & Para1 & 0.68 & 0.85 \\ \hline
        gpt4o & Para2 & 0.69 & 0.85 \\ \hline
        gpt4o & Simpa1 & 0.66 & 0.85 \\ \hline
        Qwen2.5:72b & Boukes & 0.66 & 0.85 \\ \hline
        Qwen2.5:72b & Jaidka & 0.66 & 0.85 \\ \hline
        Qwen2.5:72b & Para2 & 0.68 & 0.85 \\ \hline
        Qwen2.5:72b & Simpa1 & 0.65 & 0.85 \\ \hline
        Llama3.3:70b & Boukes & 0.73 & 0.84 \\ \hline
        Llama3.3:70b & Jaidka & 0.69 & 0.84 \\ \hline
        Llama3.3:70b & Para1 & 0.70 & 0.84 \\ \hline
        Llama3.3:70b & Para2 & 0.73 & 0.84 \\ \hline
        Llama3.3:70b & Simpa1 & 0.72 & 0.84 \\ \hline
        Llama3.1:8b & Boukes & 0.45 & 0.45 \\ \hline
        Llama3.1:8b & Jaidka & 0.45 & 0.45 \\ \hline
        Llama3.1:8b & Para1 & 0.67 & 0.79 \\ \hline
        Llama3.1:8b & Para2 & 0.55 & 0.84 \\ \hline
        Llama3.1:8b & Simpa1 & 0.6 & 0.84 \\ \hline
    \end{tabular}
\end{table}

Table~\ref{tab:perf_rationality} shows the standard performance metrics for all GLLM annotators. All annotators had a 0.84 or 0.85 accuracy score, except those using the Jaidka-prompts. There was more variation in terms of $F_1$ ranging from 0.45 (Llama3.1:8b -- Jaidka) to 0.73 (Llama3.3:70b -- Boukes), but most GLLM annotators (17/20) had an $F_1 \ge 0.6$. While this performance is certainly not perfect, they are reasonable for a difficult concept such as rationality \citep{stolwijk_can_2025}.

\begin{figure}[htbp]
  \centering
  \includegraphics[width=\linewidth,keepaspectratio]{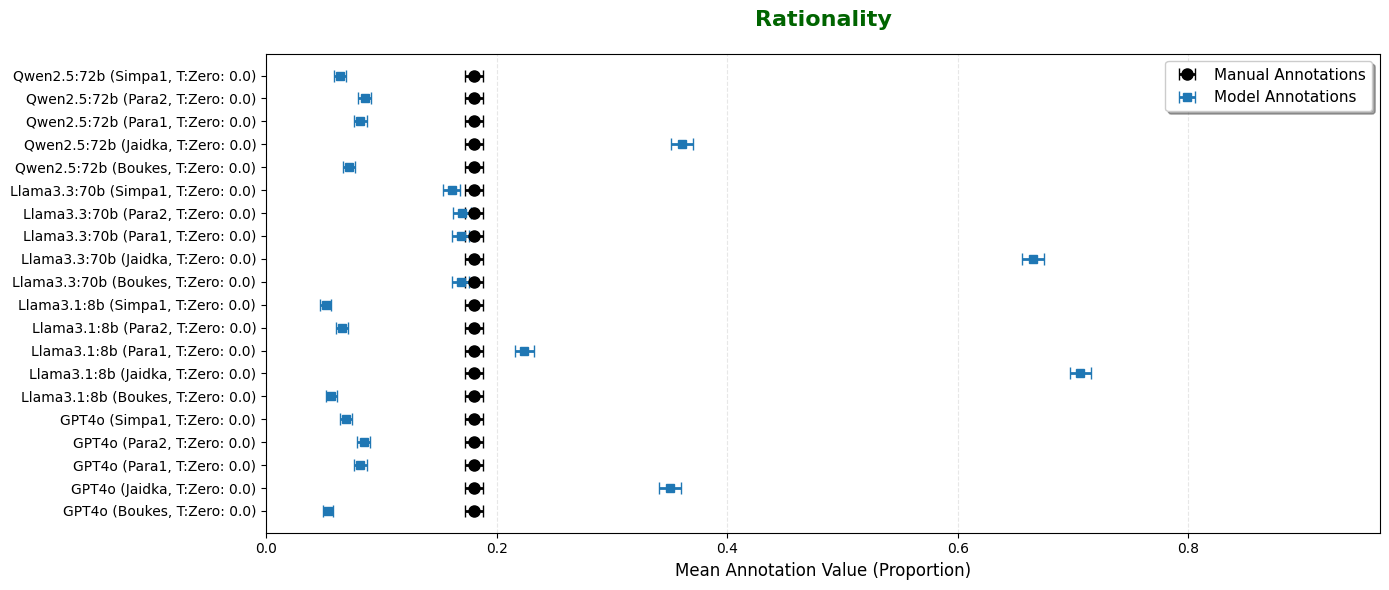}
  \caption{Estimated prevalence of rationality according to GLLM annotators compared to manual prevalence, with 95\% confidence intervals.}
  \label{fig:prev_rationality}
\end{figure}

\begin{figure}[htbp]
  \centering
  \includegraphics[width=\linewidth,keepaspectratio]{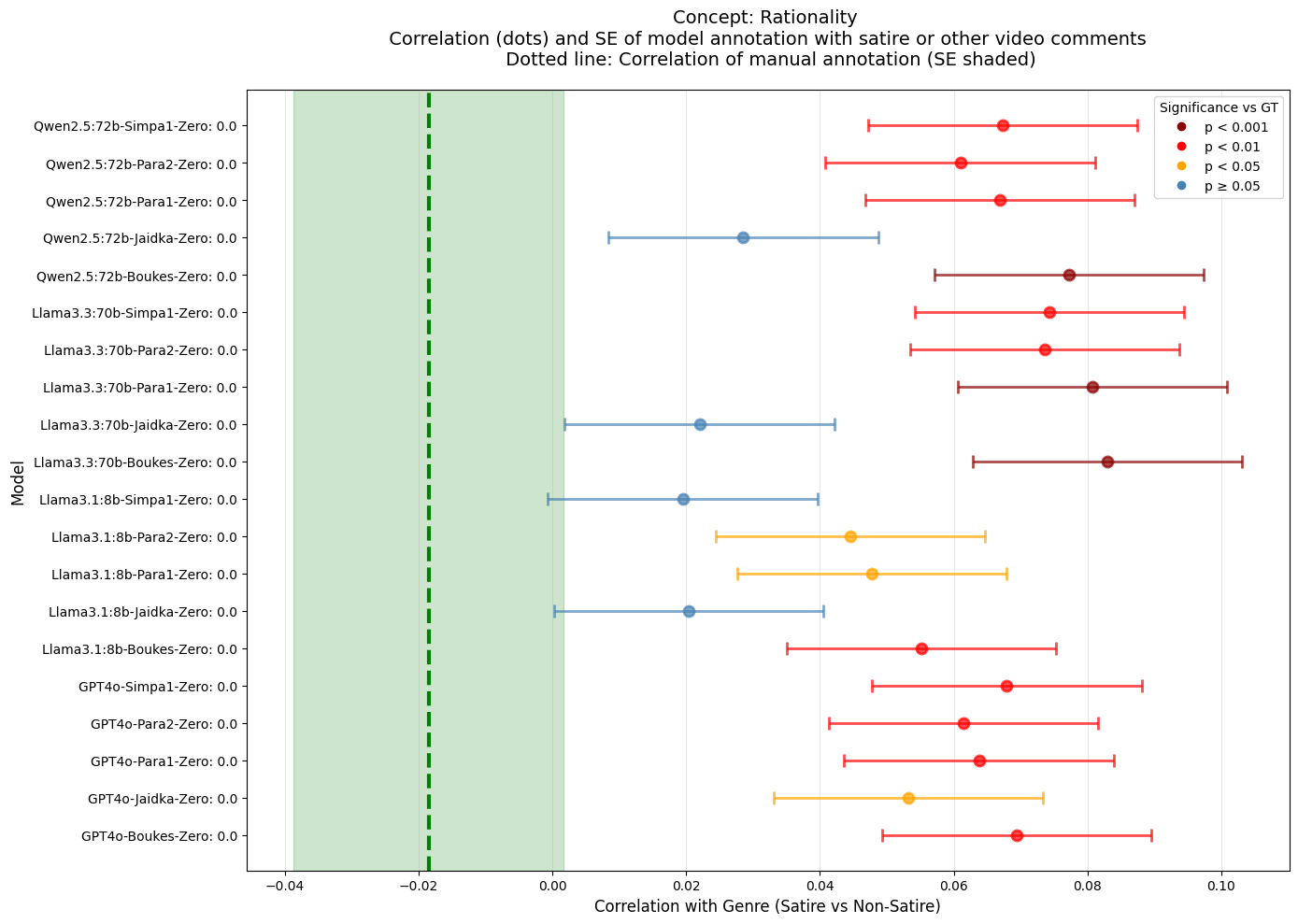}
  \caption{Estimated correlation of GLLM annotated rationality and video Genre versus manual annotated rationality.}
  \label{fig:corr_rationality}
\end{figure}

\begin{figure}[htbp]
  \centering
  \includegraphics[width=\linewidth,keepaspectratio]{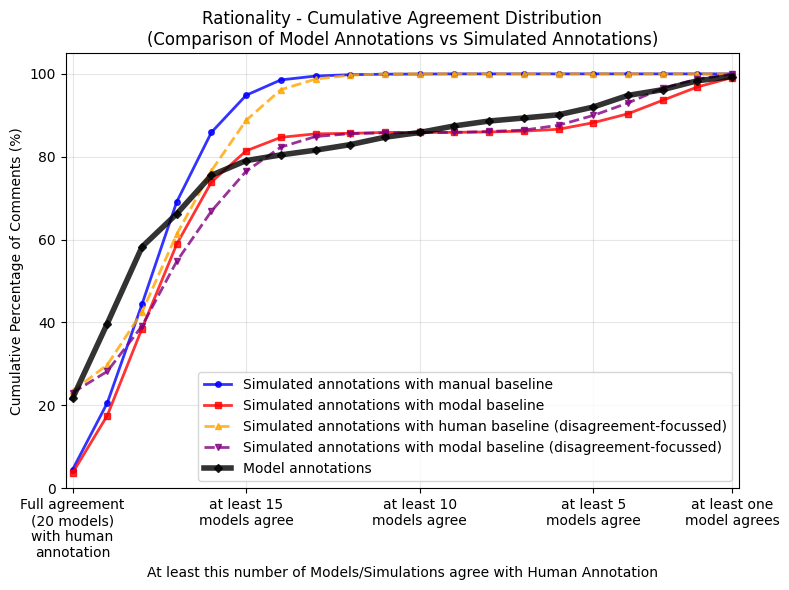}
  \caption{The number of GLLM annotators that agree with the manual annotations for what share of the YT-replies.}
  \label{fig:cumulative_agreement_rationality}
\end{figure}

\begin{figure}[htbp]
  \centering
  \includegraphics[width=\linewidth,keepaspectratio]{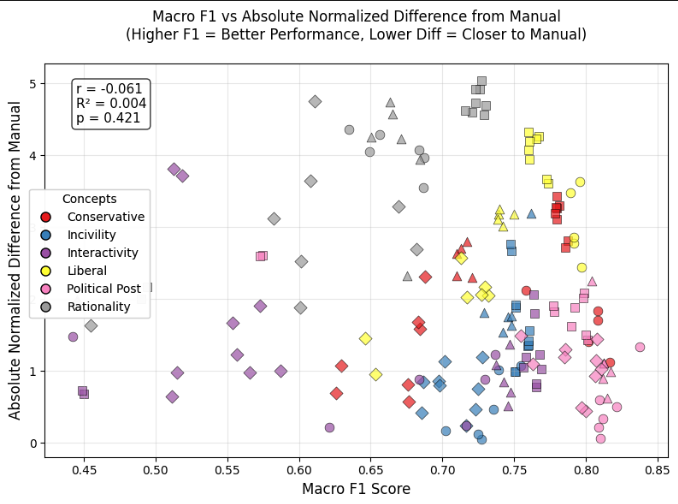}
  \caption{Bias in terms of normalized correlation coefficient difference between GLLM and manual annotation versus macro average $F_1$.}
  \label{fig:bias_vs_F1}
\end{figure}

\begin{figure}[htbp]
  \centering
  \includegraphics[width=\linewidth,keepaspectratio]{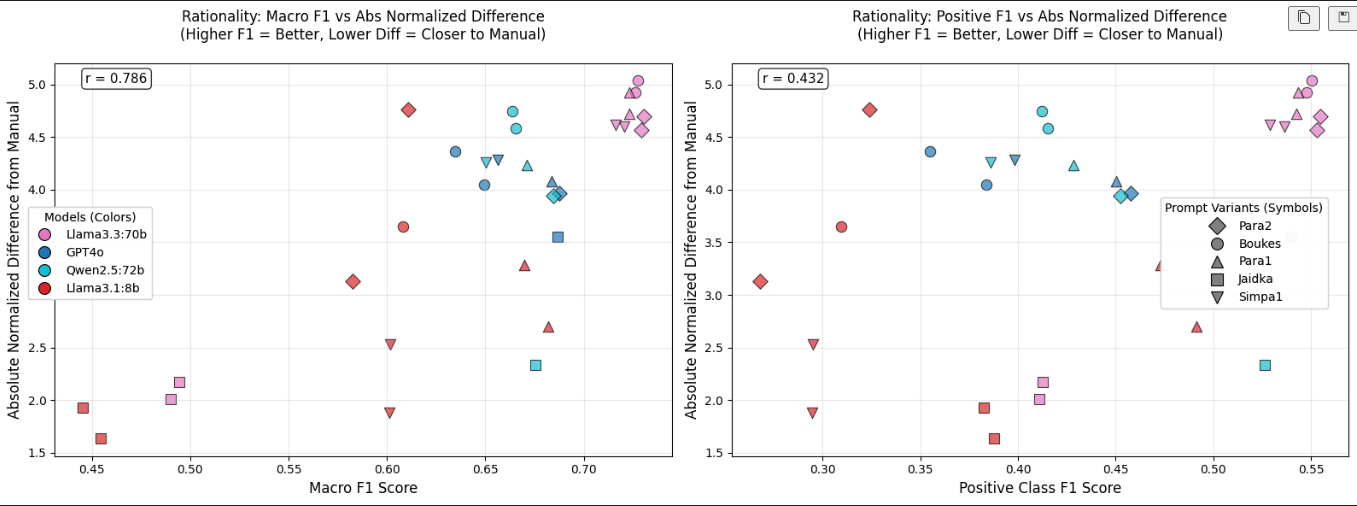}
  \caption{Bias in terms of normalized correlation coefficient difference between GLLM and manual annotation versus macro average $F_1$ and positive class $F_1$ for rationality.}
  \label{fig:bias_vs_F1_rationality}
\end{figure}

\begin{figure}[htbp]
  \centering
  \includegraphics[width=\linewidth,keepaspectratio]{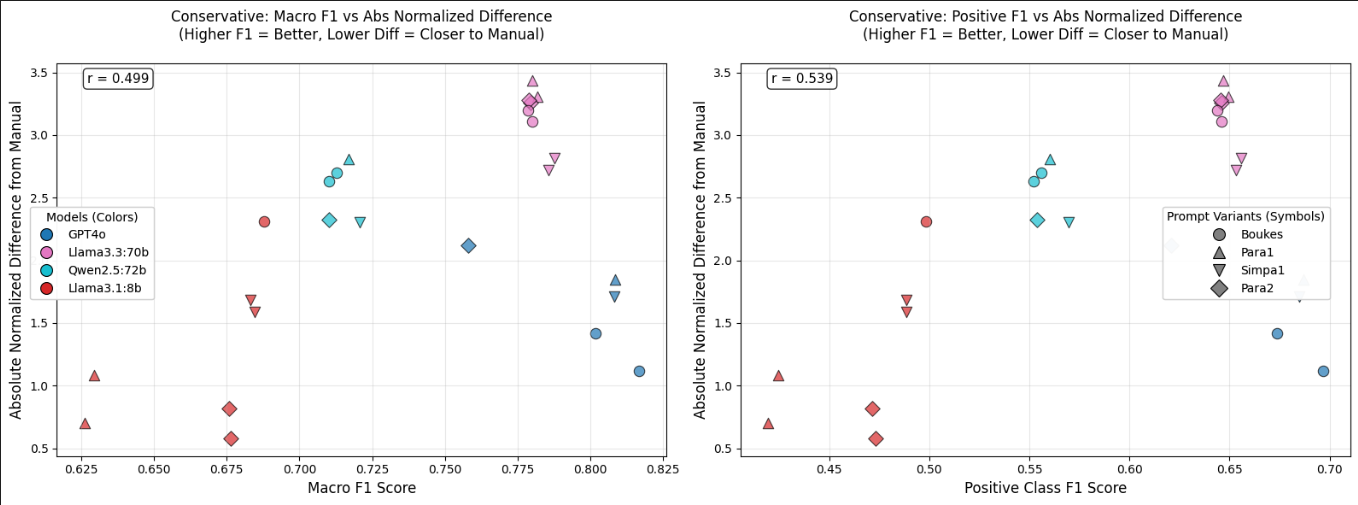}
  \caption{Bias in terms of normalized correlation coefficient difference between GLLM and manual annotation versus macro average $F_1$ and positive class $F_1$ for conservative.}
  \label{fig:bias_vs_F1_conservative}
\end{figure}

Figure~\ref{fig:prev_rationality} shows that the similarity in accuracy across the different GLLM annotators masks a strong variance in estimated prevalence, both related to the prompt and model used. The standard errors (whiskers) show that, except for four Llama3.3:70b annotators, the GLLM estimated prevalence of rationality is significantly different from the manual prevalence ($p<0.05$). The Jaidka prompts lead to more replies being classified as rational, while the different Boukes prompt-variations appear to have a much smaller effect than the choice of model, with all Boukes prompt-variations using Qwen2.5 and GPT4o underestimating the prevalence of rationality compared to both the manual annotations and those of Llama3.3:70b. 

To inspect the potential downstream effect of these differences, we plot the correlation between the amount of rational YT-replies and whether the reply was to a satire (1) or non-satire\footnote{any other genres included in the paper} (0) news video. Since our goal is to inspect GLLM biases rather than evaluate the robustness of \citet{boukes_deliberation_2024}, this simplification helps to avoid any dependencies of our results on the complex modeling setup of \citet{boukes_deliberation_2024}. Figure~\ref{fig:corr_rationality} shows that the correlation between video genre and manually coded rationality does not significantly differ from zero. However, we find a positive correlation for all GLLMs, and for most this correlation differs significantly from the correlation based on the manual annotations. Contrary to their inferior performance reported in Table~\ref{tab:perf_rationality}, the GLLMs using the Jaidka-prompt are the only ones that do \emph{not} significantly differ from the correlation coefficient found using manual annotations (green shaded area indicates the 95\% confidence interval around the correlation between the manual rationality annotations and video-genre). %These results suggest using a GLLM for annotation can indeed affect downstream results and imply that the GLLM annotations differ from the manual annotation in non-random ways, i.e. in ways related to video-genre in this case.

Figure~\ref{fig:cumulative_agreement_rationality} plots the overlap between combinations of the 20 GLLM annotators and the manual annotations. The black line shows that for about 20\% of all YT-replies, all 20 GLLM annotators arrived at the same annotation as the manual coders (Full agreement). For 40\% of all manual annotations, at least 19 out of our 20 GLLM annotators selected the same label. This percentage goes up quickly: for 80\% of the manual annotations, at least 15 out of 20 GLLMs agree. This illustrates the relatively strong performance of the GLLMs for this task and their overlap with manual annotations. 

To evaluate whether different GLLM annotations differ from manual annotations in a random or systematic fashion, we benchmark the overlap between the GLLMs with 4 sets of 20 simulated annotators (the same number as our GLLM annotators). Together these simulations help evaluate whether (1) random variation, (2) non-trivial random variation, (3) overall systematic bias or (4) non-trivial systematic bias best explains the difference between GLLM and manual annotations. We thus calculate how much overlap between manual and automatic annotations could be expected based on independent random variations. Table~\ref{tab:perf_rationality} showed that most GLLMs have an 85\% accuracy and thus differed on 15\% of annotations. Accordingly, we constructed sets of 20 simulated annotators where each annotator had a similar 85\% accuracy. 

The first set of simulated annotators added noise of 15\% randomly selected YT-replies to the manual annotations. For these selected replies, we flipped the annotation label from positive to negative or vice versa. The blue line in Figure~\ref{fig:cumulative_agreement_rationality} shows that it is statistically unlikely that random variations of 20 models would result in 20\% of YT-replies receiving the same label as manual annotations for all 20 models, like we found for the GLLMs (black line). Likewise, the agreement between the simulated annotators increases much faster than we observe between the GLLM annotations: nearly all YT-replies receive the same annotation as given by the manual annotator for at least 13 out of 20 simulated annotators, compared to only about 80\% for our GLLM annotations. Together, this shows that the GLLMs \emph{agree} more than random chance would expect for at least these 20\% of YT-replies, while they also \emph{disagree} more with the manual annotations than random chance would predict for the remaining 80\% of YT-replies. 

Perhaps there is just a set of `easy' YT-replies, which inflates the convergence between the GLLMs. The yellow line in Figure~\ref{fig:cumulative_agreement_rationality} shows overlap between another set of 20 simulated annotators, which again have an 85\% accuracy, but now only deviate from the human annotations on the presumed non-trivial YT-replies (i.e. any reply except the 20\% with full agreement). The yellow line now coincides with the black line for full agreement of all 20 simulated annotators, but again increases much faster and starts overlapping with the blue line since they likewise agree with all human annotations for at least 13 simulated annotators or more. This shows that GLLMs still have a stronger than random overlap for the remaining 80\% of YT-replies than one would expect by chance. 

The red and purple line in Figure~\ref{fig:cumulative_agreement_rationality} use a similar approach with two new sets of simulations, but they assume the modal annotation of the GLLMs to be the ground truth to which random deviations are added. The thus assume a systematic bias: GLLMs agree more with each other than with the manual annotations. These simulated annotators thus have an 85\% accuracy with the modal GLLM annotation. The red line shows that regardless of baseline (human or modal GLLM annotations), the chances of agreeing with human annotations for all 20 annotators simultaneously, based on random variations, is negligible, and much lower than observed among our GLLM annotators. The purple line shows that random deviations from the modal GLLM annotation, excluding the 20\% 'easy' YT-replies, best explains the data, suggesting that there is indeed a common core to how GLLM annotations deviate from the manual annotations: They consistently agree in their classification of the 20\% easy YT-replies, but agree more with each other than with the manual annotations for the remaining 80\%, suggesting a systematic bias.

In practice, however, we would first run a set of prompts and models to evaluate their performance and only proceed with the best performing model for our downstream task. Although $F_1$ is not designed to detect bias, the best performing model has a better overlap with manual annotations to begin with, arguably decreasing the space for and size of bias. To test whether such standard procedure helps to minimize bias, we again use the correlation coefficient described above (plotted for rationality in Figure~\ref{fig:corr_rationality}), but now for all concepts. We calculated the bias of each model in terms of the difference of this correlation with the same correlation based on manual annotations (normalized to ensure comparability across concepts). Figure~\ref{fig:bias_vs_F1} plots the relation between the $F_1$ score of each GLLM-prompt combination and its bias. The plot shows no significant correlation between the bias and $F_1$ score, suggesting that selecting on $F_1$ is not beneficial to reduce bias. However, if we take a closer look at each concept (i.e. color in ~\ref{fig:bias_vs_F1}) individually, we appear to observe a positive correlation. This suggests that bias increases with $F_1$. Figure~\ref{fig:bias_vs_F1_rationality} illustrates this for rationality, both in terms of macro average $F_1$ and positive class $F_1$ (the performance in correctly annotating a comment as rational). Both metrics are positively related to the bias of the GLLM annotator. This means that selecting the better performing GLLM in terms of $F_1$ score would \emph{increase} rather than decrease the difference in result on the downstream task compared to manual annotations.

\subsection*{Discussion and conclusion}
Our results show that the choice of GLLM and the prompt wording can influence the resulting annotations, both in the prevalence of a concept and its substantial meaning (bias). The positive relation between bias and $F_1$-score shows that these results cannot be explained based on the noise in the manual annotations themselves, since these metrics already include this noise, and we find similar results for 'easier' concepts like whether a comment is conservative, see Figure~\ref{fig:bias_vs_F1_conservative}. Both \citet{egami_using_2024} and \citet{angelopoulos_prediction-powered_2023} present ways to address biased GLLM annotations by combining them with manual annotations, but only work for small to medium size samples. Therefore, our results caution against using generative AI models for large-scale text annotation tasks without evaluating whether the downstream results depend on the chosen annotation model. Furthermore, traditional performance metrics failed to detect bias. We recommend further research to propose more bias-sensitive metrics.

\subsection*{Acknowledgments}
An earlier version of this paper was presented at IC2S2 2025. We thank all participants for their inputs that helped improve this paper.

%\section*{References}
\bibliographystyle{plainnat}
\bibliography{references}
%\printbibliography % Print the bibliography

\end{document}